 \documentclass[final,5p,times,twocolumn]{elsarticle}

\usepackage{amssymb}
\usepackage{amsthm}
\usepackage{array}
\usepackage{times}
\usepackage{epsfig}
\usepackage{graphicx}
\usepackage{amsmath}
\usepackage{amssymb}
\usepackage{epstopdf}
\usepackage{url}
\usepackage{comment}
\usepackage{subfigure}
\usepackage[font=small,labelfont=bf]{caption}
\usepackage{lineno}

\newcommand{\PreserveBackslash}[1]{\let\temp=\\#1\let\\=\temp}
\newcolumntype{C}[1]{>{\PreserveBackslash\centering}p{#1}}
\newcolumntype{R}[1]{>{\PreserveBackslash\raggedleft}p{#1}}
\newcolumntype{L}[1]{>{\PreserveBackslash\raggedright}p{#1}}

\journal{Neurocomputing}

\begin{document}

\begin{frontmatter}



\title{Video Interpolation using Optical Flow and Laplacian Smoothness}


\author[label1,label2]{Wenbin Li}
\author[label2]{Darren Cosker}

\address[label1]{Department of Computer Science, University College London, UK}
\address[label2]{Centre for the Analysis of Motion, Entertainment Research and Applications (CAMERA), University of Bath, UK}

\begin{abstract}

Non-rigid video interpolation is a common computer vision task. In this paper we present an optical flow approach which adopts a Laplacian Cotangent Mesh constraint to enhance the local smoothness. Similar to Li~\emph{et al.}, our approach adopts a mesh to the image with a resolution up to one vertex per pixel and uses angle constraints to ensure sensible local deformations between image pairs. The Laplacian Mesh constraints are expressed wholly inside the optical flow optimization, and can be applied in a straightforward manner to a wide range of image tracking and registration problems. We evaluate our approach by testing on several benchmark datasets, including the Middlebury and Garg~\emph{et al.} datasets. In addition, we show application of our method for constructing 3D Morphable Facial Models from dynamic 3D data.

\end{abstract}


\end{frontmatter}


\section{Introduction}

Non-rigid video interpolation is a computer vision related problem that requires the tracking of non-rigid objects, calculation of dense image correspondences and the registration of image sequences containing highly non-rigid deformation. Existing algorithms to achieve this include model based tracking~\cite{AAMs}, dense patch identification and matching~\cite{Dense-patches,Pizarro}, group-wise image registration~\cite{groupwise}, space-time tracking~\cite{space-time,APO,APO_JIFS,moBlur,moblur_nc} and optical flow~\cite{HS,Brox,Garg,tang,gt,li2013nonrigid}. All such models and the general dense tracking have been widely used in fields e.g. motion tracking~\cite{lv2014multimodal,lv2015touch}, visualization~\cite{lv2013game,su2016multi} and interaction~\cite{tv}.

Optical flow is an attractive formulation as it provides a dense displacement field between image pairs. In most standard approaches, assumptions regarding gray value constancy between images and smoothness in motion between neighboring pixels are adopted~\cite{Brox,HS}. Sun~\emph{et al.}~\cite{Sun} propose a different approach which overcome these constraints by learning a probabilistic model for flow estimation. However, their approach requires training pre-calculated optical flow ground truths, which are difficult to obtain. In the general optical flow model, it is common to adopt a data term consisting of gray value and gradient constraints (e.g. Brox and Malik~\cite{Brox}) and an additional smoothness term. Nevertheless, most previous optical flow formulations only consider global smoothness and ignore formulations that preserve local image details.

Many optical flow techniques concentrate on problems where the scene movement is largely rigid in nature. However, there are many problem cases where we would like to calculate flow given highly non-rigid global and local image displacements over long image sequences. One recent problem highlighting this particular case is the alignment of 3D dynamic facial sequences containing highly non-rigid deformations~\cite{cosker,Bradley}. The problem requires non-rigidly aligning a set of images to a reference - e.g. a neutral facial expression. Each image referred to as a UV map~\footnote{UV refers to the XY location of a pixel in the image. UV map is the graphical term for the \emph{texture} for a 3D model. Each UV location maps to a 3D vertex on a corresponding mesh} is accompanied by a corresponding 3D mesh, and each mesh has a difference vertex topology. Once the UV maps are registered to a reference image (e.g. a neutral expression), vertex correspondence can be imposed. The technique is popular in 3D Morphable Model construction~\cite{cosker,Blanz}.

Beeler et al~\cite{Beeler} and Bradley et al~\cite{Bradley} adopt a slightly different approach to mesh correspondence. In their solutions, image displacement is calculated from camera views and then used to deform a reference mesh from an initial frame through a 3D sequence. The optical flow provides guides for adjusting pixel positions, and the mesh reduces artefacts by imposing a constraint to prevent faces on the mesh from becoming inverted or flipped. Further, mesh and image deformation research in graphics is an active area of research~\cite{BBW}. Such techniques provide flexible methods to invoke deformation while preserving some desired properties such as local geometric details. As such, it is also of interests to apply such solutions as smoothness constraints to optical flow calculation, which forms the central basis of our presented work. Li~\emph{et al.}~\cite{LME} introduce a hybrid optical flow framework that takes into account a laplacian mesh data term and a global smoothness term. However, their energy is highly nonlinear and hard to minimize.

\subsection{Contributions}

In this paper we present an optical flow algorithm (\emph{LCM-flow}) which adopts a smoothness term based on \emph{Laplacian Cotangent Mesh Deformation}. Such deformation approaches has been widely used in graphics, particularly for preserving small details on deformable surface~\cite{sorkine2005laplacian,meyer2002discrete}. Such concept shows advantage in the non-rigid optical flow estimation~\cite{LME}. Those energy is able to penalizes local movements and preserves smooth global details. In our method, the proposed constraint on the local deformations is expressed in Laplacian coordinates encourage local regularity of the mesh whilst allowing globally non-rigid preservation.

Similar to Li~\emph{et al.}~\cite{LME}, our proposed algorithm applies a mesh to the image with a resolution up to one vertex per pixel. The Laplacian constraint is described in terms of a smoothness term, and can be applied in a straightforward manner to a number of optical flow approaches with the addition of our proposed minimization strategy. We evaluate our approach on the popular \emph{Middlebury} dataset~\cite{Middlebury} as well as the publicly available non-rigid data set proposed by Garg~\emph{et al.}~\cite{Garg}. We show our method to give high performance on \emph{Middlebury} in terms of interpolation, and either outperform or show comparable accuracy against the leading publicly available non-rigid approaches when evaluated against Garg~\emph{et al.} In addition, we show an application of our optical flow approach for building dynamic \emph{3D Morphable Models} from dynamic 3D facial data, and outperform a current state of the art method.

The remainder of our paper is organized as follows: In sections~\ref{sec:EnergyFunction} and \ref{sec:Minimisation} our strategy for calculating optical flow displacements between image pairs is outlined. Section~\ref{sec:evaluation} shows an evaluation of \emph{LCM-flow} on the \emph{Middlebury} data set and four other publicly available sequences of non-rigidly deforming objects~\cite{Garg}.

\vspace{-2mm}

\section{Energy Function Definition}
\label{sec:EnergyFunction}

In this section the core energy function of our \emph{Laplacian Cotangent Mesh} based Optical Flow approach is presented. In the formulation the algorithm considers a pair of consecutive frames in an image sequence. The current frame is denoted by $I_{i}(\textbf{X})$ and its successor by $I_{i+1}(\textbf{X})$, where $\textbf{X} = (x,y)^{T}$ is a pixel location in the image domain $\Omega$. We define the optical flow displacement between $I_{i}(\textbf{X})$ and $I_{i+1}(\textbf{X})$ as $\textbf{w}_{i} = (\textbf{u},\textbf{v})^{T}$. In the proposed optical flow estimation approach, the core energy function can be obtained from the following general formulation:

\begin{align}
\textbf{E}(\textbf{w}) = \textbf{E}_{Data}(\textbf{w}) + \lambda\cdot\textbf{E}_{Global}(\textbf{w}) + \xi\cdot\textbf{E}_{Lap}(\textbf{w}) \nonumber
\end{align}

where $\textbf{E}_{Data}(\textbf{w})$ denotes a data term that contains both \emph{Gray Value} and \emph{Gradient Constancy} assumptions (Section~\ref{sec:Edata}) on pixel values between $I_{i}(\textbf{X})$ and $I_{i+1}(\textbf{X})$.

Two smoothness terms are also introduced into the formulation. Similar to ~\cite{Brox,HS}, the first term $\textbf{E}_{Global}(\textbf{w})$ controls global flow smoothness. The second term represents our core contribution, i.e. a \emph{Laplacian Cotangent Mesh} constraint $\textbf{E}_{Lap}(\textbf{w})$. In the following sections we next describe each term in detail, focusing on our Laplacian constraint in Section 2.3.

\vspace{-2mm}

\subsection{Data Term Definition}
\label{sec:Edata}

Following the standard optical flow assumption regarding \emph{Gray Value Constancy}, we assume that the gray value of a pixel is not varied by its displacement through the entire image sequence. In addition, we also make a \emph{Gradient Constancy} assumption which is engaged to provide additional stability in case the first assumption (\emph{Gray Value Constancy}) is violated by changes in illumination. The data term of \emph{LCM-flow} encoding these assumptions is therefore formulated as:

\begin{align}
\textbf{E}_{Data}(\textbf{w}) &= \sum_{\Omega}  \Psi  (I_{i+1}(\textbf{X}+\textbf{w}) - I_{i}(\textbf{X}))^{2} \nonumber \\
&+ \theta\cdot \sum_{\Omega} \Psi (\nabla I_{i+1}(\textbf{X}+\textbf{w}) - \nabla I_{i}(\textbf{X}))^{2} \nonumber
\end{align}

In order to deal with occlusions, we apply the increasing concave function $\Psi(s^{2}) = \sqrt{s^{2} + \epsilon^{2}}$ with $\epsilon = 0.001$~\cite{Brox} to solve this formation which enables $L1$ minimization. The remaining term $\nabla = (\partial_{xx},\partial_{yy})^{T}$ is a spatial gradient and $\theta\in[0,1]$ denotes weight that can be manually assigned with different values. In the experiments it is pre-defined as 0.5.

\vspace{-2mm}

\subsection{Global Smoothness Constraint}
\label{sec:gSmooth}

The first smoothness term of \emph{LCM-flow} is a dense pixel based regularizer that penalizes global variation. The objective is to produce a globally smooth optical flow field (
as in the data term, the robust function $\Psi(s^{2})$ is again used):

\begin{align}
\textbf{E}_{Global}(\textbf{w}) = \sum_{\Omega} \Psi(\mid \nabla \textbf{u} \mid^{2} + \mid \nabla \textbf{v} \mid^{2}) \nonumber
\end{align}


\subsection{Laplacian Cotangent Mesh Smoothness Constraint}
\label{sec:LCMterm}


Global smoothness terms are widely used in optical flow formulation~\cite{Middlebury}. However, their definition means that local nonlinear variations between images - such as those in non-rigid motion - can be over smoothed. In order to improve optical flow estimation against the local complexity of non-rigid motion, a novel \emph{Laplacian Cotangent Mesh} constraint is proposed in this section. The aim of this constraint is to account for non-rigid motion in scene deformation. This term is inspired by Laplacian mesh deformation research in graphics which aims to preserve local mesh smoothness under non-linear transformation~\cite{sorkine2005laplacian}. It's use in computer vision research for optical flow estimation is introduced for the first time here. Although non-rigid motion is highly nonlinear, the movement of pixels in such deformations still often exhibit strong correlations in local regions. In order to represent this, we propose a quantitative \emph{Cotangent Weight} based on a Laplacian framework and a differential representation. The scheme was originally presented by Meyer~\emph{et al.}~\cite{meyer2002discrete} for mesh deformation.


\begin{figure*}[t]
\centerline{
\includegraphics[width=0.95\linewidth]{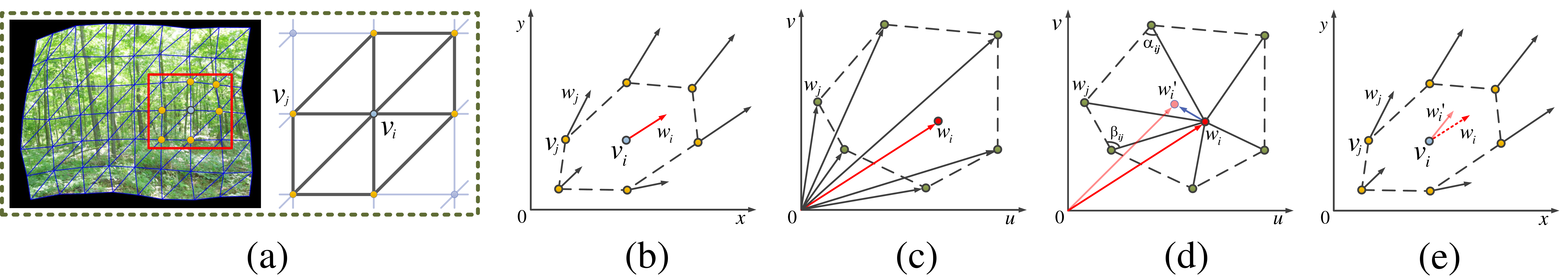}
}
\caption{Laplacian cotangent mesh constraint in optical flow vector space. (a): A mesh on a specific frame. (b): 1-ring neighborhood based on vertices. (c): 1-ring neighborhood based on endpoints of optical flow vectors. (d): $\delta(\textbf{w}_i)$ (the blue vector) calculated by endpoints of optical flow vectors. (e): The modified optical flow vector $\textbf{w}'_i$ based on $\textbf{w}_i$ and $\delta(\textbf{w}_i)$.}
\label{fig:triFlow}
\end{figure*}

We assume that the image is initially covered by a triangular mesh denoted by $M = (V,E,F)$, where $n$ is the number of vertices, $V$ is the set of vertex coordinates, $E$ is the set of edges, and $F$ is the set of faces. The location of each vertex is represented by absolute cartesian coordinates. The $i$-th vertex is denoted by $\textbf{v}_{i} \in V$. Considering a small mesh region, each vertex $\textbf{v}_{i}$ has a 1-ring neighborhood denoted by $\mathcal{A}_{ring}$, as shown in Figure~\ref{fig:triFlow}. The motion relationship between vertex $\textbf{v}_{i}$ and its adjacent neighbors in $\mathcal{A}_{ring}$ can be represented as follows:

\begin{align}
\delta(\textbf{v}_{i}) = \sum_{j \in N(i)} \frac{1}{2}(cot \alpha_{ij} + cot \beta_{ij})(\textbf{v}_{i} - \textbf{v}_{j})
\label{eq:LCMorig}
\end{align}

where $N(i) = \lbrace j \mid (i,j) \in E \rbrace$, $\alpha_{ij}$ and $\beta_{ij}$ are the two angles opposite the edge $\textbf{e}_{ij} = \textbf{v}_{i} - \textbf{v}_{j}$. In a planar region, every vertex $\textbf{v}_{i}$ in a small region of an individual object is assumed to have its motion trend towards a new position. This trend can be quantified by the vector $\delta(\textbf{v}_{i})$. This is a core concept for Laplacian based mesh deformation. It is adopted widely in computer graphics because it enables mesh deformations that preserve the local surface shape whilst allowing significant global motion.~\cite{sorkine2005laplacian}.

We assume that this technique can be applied as constraints in optical flow vector space. Figure~\ref{fig:triFlow}(a-e) illustrates the core steps of this assumption. However, the original equation~\eqref{eq:LCMorig} omits the effect of the 1-ring neighborhood which in previous work has been shown to give additional deformation robustness~\cite{meyer2002discrete}. Therefore, the \emph{Voronoi Region Area} of $\textbf{v}_{i}$ denoted by $\mathcal{A}_{Voronoi}^i$ is introduced to the formulation. This provides additional robustness given different dimensions of $\mathcal{A}_{ring}$. We have

\begin{align}
\delta(\textbf{w}) &= \sum_{i \in V}[\frac{\sum_{j \in N(i)}(cot \alpha_{ij} + cot \beta_{ij})(\textbf{w}_{i} - \textbf{w}_{j})}{ 2\mathcal{A}_{Voronoi}^{i}}] \\
\mathcal{A}_{Voronoi}^{i} &= \frac{1}{8}\sum_{j \epsilon N(i)} (cot \alpha_{ij} + cot \beta_{ij}) |  \textbf{w}_{i} - \textbf{w}_{j}|^{2}
\label{fig:LCMvector}
\end{align}

The motion trend of the optical flow vector can be presented by (2) and (3) and are introduced as the core contribution to our energy formulation. Where $\delta(\textbf{w}) = (\delta_{u}, \delta_{v})^{T} $, $\delta_{u}$ is motion trend in the horizontal direction and $\delta_{v}$ motion trend in the vertical direction. In addition, $\textbf{w}_*$ is the optical flow vector at the position of vertex $\textbf{v}_*$. Based on these formulations, the high order smoothness constraint based on \emph{Laplacian Cotangent Mesh} is defined as:

\begin{align}
E_{Lap}(\textbf{w}) = \sum_{n(V)} \Psi(|\nabla \delta_{u}|^{2} + |\nabla \delta_{v}|^{2})
\end{align}

In this formulation, the \emph{Laplacian Cotangent Mesh} constraint helps greatly to preserve local details by minimizing the angle differences -- or preserving the geometric details as in Laplacian mesh deformation~\cite{sorkine2005laplacian} -- of local neighborhoods. The main reason for this is that by constraining local edge angles -- embedded in the \emph{Laplacian Cotangent Mesh} constraint formulation -- the term penalizes local displacements which may cause overlaps with other pixels. The experiments (Figure~\ref{fig:triPer}) in Section~\ref{sec:evaluation} demonstrate how the constraint achieves the better preservation of image details given non-rigid motion changes.

\vspace{-2mm}

\section{Minimization}
\label{sec:Minimisation}


Due to the highly non-linear nature of the energy function $\textbf{E}(\textbf{w})$, our minimization approach is an essential part in our work. We apply multiple nested fixed point iterations to minimize our energy function that includes our novel \emph{LCM} smoothness term. In this section, we introduce the numerical minimization scheme for our \emph{LCM} smoothness term. This includes a discrete computation scheme for our Laplacian operator and the gradient magnitude on $\delta_{*}$ (see section~\ref{sec:LO_GM}). We initially define mathematical abbreviations for our minimization as follows (using the same notation as in ~\cite{Brox}):

\begin{align}
\begin{array}{ll}
I_{x}=\partial_{x}I(\textbf{X}+\textbf{w})& I_{yy}=\partial_{yy}I(\textbf{X}+\textbf{w}) \\
I_{y}=\partial_{y}I(\textbf{X}+\textbf{w}) & I_{xx}=\partial_{xx}I(\textbf{X}+\textbf{w}) \\
I_{z}=I_{i+1}(\textbf{X}+\textbf{w})-I_{i}(\textbf{X}) & I_{xz}=\partial_{x}I_{i+1}(\textbf{X}+\textbf{w})-\partial_{x}I_{i}(\textbf{X})\\
I_{xy}=\partial_{xy}I(\textbf{X}+\textbf{w}) & I_{yz}=\partial_{y}I_{i+1}(\textbf{X}+\textbf{w})-\partial_{y}I_{i}(\textbf{X}) \\
\end{array}
\end{align}

\subsection{First Fixed Point Iterations}

In order to minimize the energy function $\textbf{E}(\textbf{w})$, the Euler-Lagrange equations can be employed. However, the Euler-Lagrange equations obtained are still highly nonlinear with the argument $\textbf{w}=(u,v)^T$. In order to solve this system, we apply two nested steps of \emph{Fixed Point Iterations} on $\textbf{w}$. Before the first fixed point iterations, a down sampling image pyramid is constructed. The algorithm goes through every level of the pyramid and starts on the top/coarsest level. For each level of the pyramid, we assume that $\textbf{w}$ converges at the $k$-th iteration giving us $\textbf{w}^k=(\textbf{u}^k,\textbf{v}^k)^T, k=0,1,\dots$ with the initialization $\textbf{w}^0 = (0,0)^T$ at the coarsest level of the pyramid. The flow field $\textbf{w}^k$ is then to be propagated as the initial flow field for $\textbf{w}^{k+1}$ on the next finer level of the pyramid.

\begin{figure*}[t]
	\centering
	\includegraphics[width = 0.95\textwidth]{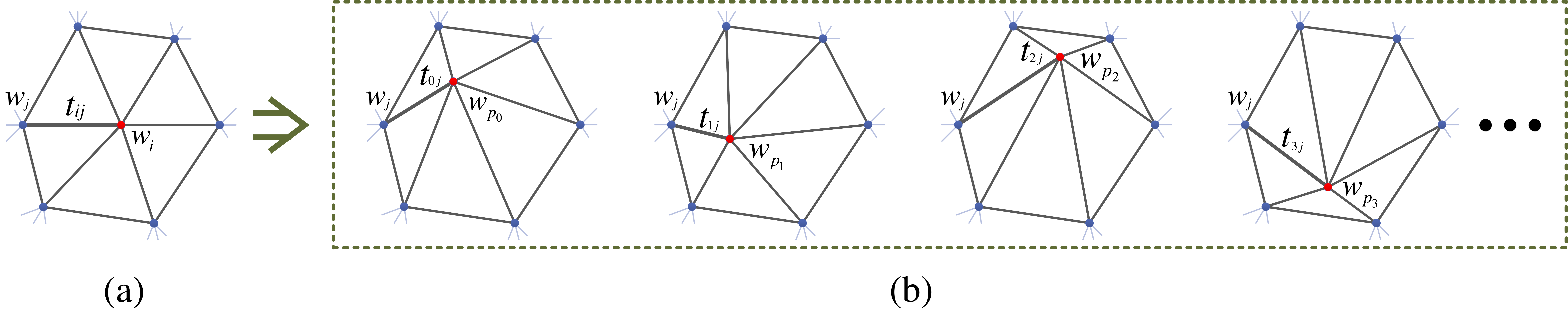}
    \caption{Computing $\delta_*$ within 1-ring neighborhood. (a): Weight $t_*$ in computation of $\overline{\delta_*^{i}}$. (b): Our optimization traverses every pixel and compute the $\delta_*$ within their 1-ring neighborhood. Those neighborhoods are preset by the initial mesh structure.}
	\label{fig:devi}
\end{figure*}

\subsubsection{$\delta_{*}$ Vector Field Computation}

The $\delta_{*}$ vector field is updated during every step of the first fixed point iteration. According to the formulation in section~\ref{sec:LCMterm}, the $\delta_{u}(u^{k+1})$ can be calculated from:

\begin{align}
\delta_{u}(u^{k+1}) &= \sum_{i \in V}[\frac{\sum_{j \in N(i)}(cot \alpha_{ij} + cot \beta_{ij})(u_{i}^{k+1} - u_{j}^{k+1})}{2\mathcal{A}_{Voronoi}^i}] \\
\mathcal{A}_{Voronoi}^{i} &= \frac{1}{8}\sum_{j \epsilon N(i)} (cot \alpha_{ij} + cot \beta_{ij}) |  \textbf{w}_{i}^{k+1} - \textbf{w}_{j}^{k+1}|^{2}
\end{align}

$\delta_{v}(v^{k+1})$ is computed in a similar formula, where $\textbf{w}_*^{k+1}=(u_*^{k+1},v_*^{k+1})^T$ is optical flow vector at vertex $\textbf{v}_*$. $|\textbf{w}_{i}^{k+1} - \textbf{w}_{j}^{k+1}|$ is \emph{Euclidean Distance} between the endpoints of $\textbf{w}_{i}^{k+1}$ and $\textbf{w}_{j}^{k+1}$. From perspective of implementation, an individual loop is necessary for calculating $\delta_{*}$ from the information of every angle within each iteration step. As shown in Figure~\ref{fig:devi}, $\delta_*$ is calculated on every pixel within the same neighborhood based on the initial mesh structure.

\subsection{Nested Second Fixed Point Iterations}

After the first fixed point iterations, the new system of equations is still nonlinear and difficult to solve due to this contains terms of $I_{*}^{k+1}$ and the nonlinear functions $\Psi'$. First order Taylor expansions is employed on the nonlinear terms $I_{z}^{k+1}$, $I_{xz}^{k+1}$ and $I_{yz}^{k+1}$ to remove nonlinear terms of $I_{*}^{k+1}$. Also, we assume that $u^{k+1}=u^k+du^k$ and $v^{k+1}=v^k+dv^k$ can obtain two unknown increments $du^k$, $dv^k$ and two known flow fields $u^k$ ,$v^k$ from the previous iteration. Furthermore, in order to remove nonlinearity existed in $\Psi'_{*}$ with unknown increments $du^k$ and $dv^k$, we apply a nested second fixed point iteration. In every iteration step of second fixed point iterations, we assume that both $du^{k,l}$ and $dv^{k,l}$ converges by $l$ iteration steps with initialization of $du^{k,0}=0$ and $dv^{k,0}=0$. Therefore, the final linear system is obtained in $du^{k,l+1}$ and $dv^{k,l+1}$ as follows:

\begin{align}
(\Psi')_{Data}^{k,l}\cdot(I_{x}^k(I_{z}^k+I_{x}^kdu^{k,l}+I_{y}^kdv^{k,l+1})& \nonumber \\ +\theta~[I_{xx}^k(I_{xz}^k+I_{xx}^kdu^{k,l}+I_{xy}^kdv^{k,l+1})& \nonumber\\
+I_{xy}^k(I_{yz}^k+I_{xy}^kdu^{k,l}+I_{yy}^kdv^{k,l+1})])& \nonumber \\
-\lambda~\mathbf{Div}(\Psi')_{Global}^{k,l}\cdot\nabla(u^k+du^{k,l+1})& \nonumber \\
-\xi~\mathbf{Div}(\Psi')_{Lap}^{k,l}\cdot\nabla\delta_{u}(u^k+du^{k,l+1})&=0
\label{eq:EnergyKL_1}
\end{align}
\begin{align}
(\Psi')_{Data}^{k,l}\cdot(I_{y}^k(I_{z}^k+I_{x}^kdu^{k,l}+I_{y}^kdv^{k,l+1})& \nonumber \\ +\theta~[I_{yy}^k(I_{yz}^k+I_{xy}^kdu^{k,l}+I_{yy}^kdv^{k,l+1})& \nonumber \\
+I_{xy}^k(I_{xz}^k+I_{xx}^kdu^{k,l}+I_{xy}^kdv^{k,l+1})])& \nonumber \\
-\lambda~\mathbf{Div}(\Psi')_{Global}^{k,l}\cdot\nabla(v^k+dv^{k,l+1})& \nonumber \\
-\xi~\mathbf{Div}(\Psi')_{Lap}^{k,l}\cdot\nabla\delta_{v}(v^k+dv^{k,l+1})& = 0
\label{eq:EnergyKL_2}
\end{align}

Where $(\Psi')_{Data}^{k}$ provides robustness against occlusion, $(\Psi')_{Global}^k$ and $(\Psi')_{Lap}^k$ are defined as diffusivity in both smoothness terms \cite{brox2004high} as follows:

\begin{alignat}{1}
(\Psi')_{Data}^{k}&=\Psi'((I_{z}^k+I_{x}^kdu^k+I_{y}^kdv^k)^2 \nonumber \\
&+\theta[(I_{xz}^k+I_{xx}^kdu^k+I_{xy}^kdv^k)^2 \nonumber \\
&+(I_{yz}^k+I_{xy}^kdu^k+I_{yy}^kdv^k)^2]) \nonumber \\
(\Psi')_{Global}^{k}&=\Psi'(|\nabla(u^k+du^k)|^2+|\nabla(v^k+dv^k)|^2) \nonumber \\
(\Psi')_{Lap}^{k}&=\Psi'(|\nabla\delta_{u} (u^{k}+du^k)|^2+|\nabla\delta_{v} (v^{k}+dv^k)|^2)
\end{alignat}

\subsection{Laplacian Operator and Gradient Magnitude}
\label{sec:LO_GM}

In order to compute \emph{Div} term that refers to $(\Psi')_{Global}^{k,l}$, we have to calculate Laplacian operator and the gradient magnitudes of $|\nabla u|$ and $|\nabla v|$ in image space. Laplacian operator is practically approximated numerically based on finite differences in discrete cases. Hence we have $\nabla u = \overline{u}-u$ and $\nabla v = \overline{v}-v$, where $\overline{u}$ and $\overline{v}$ are weighted average of $u$ or $v$ and calculated by adjacent neighborhood around a specific pixel. The methods to determine $|\nabla u|$ and $|\nabla v|$ have been discussed for many years, finite differences in Faisal and Barron's work~\cite{faisal2007high} is applied in our approach.

For the second \emph{Div} term referred to $(\Psi')_{Lap}^{k,l}$, we also have to calculate Laplacian operator and the gradient magnitudes in mesh space. Note that the algorithm is proposed to adopt general mesh, each vertex of which has 6 adjacent neighbors in its 1-ring surroundings. In the implementation, we adopt the meshing step from~\cite{LME} to initialize the mesh from the input frame with flexible vertex density control. In optical flow vector space (Figure~\ref{fig:triFlow}(c-d)), we have $\nabla \delta_*$ approximated from $\nabla \delta_*^{i}=\overline{\delta_*^{i}}-\delta_*^{i}$, where $\overline{\delta_*^{i}}$  are weighted average of $\delta_*^i$ and calculated by adjacent neighborhood around a specific optical flow vector $\textbf{w}_i$. Also, the weight $t_*$ is linear and based on the absolute endpoint distance between $\textbf{w}_i$ and the adjacent neighbor (Figure~\ref{fig:devi}(a)).

After obtaining Laplacian operator and the gradient magnitudes, the linear system can be solved by using common numerical methods such as \emph{Gauss-Seidel} and \empty{Successive Over Relaxation} (\emph{SOR}). In the following section, more details about implementation will be presented.

\vspace{-2mm}

\section{Evaluation}
\label{sec:evaluation}

In this section we evaluate the accuracy of our \emph{LCM-flow} method. We also explore the behavior of our \emph{Laplacian Cotangent Mesh} (LCM) constraint by considering examples of image registration using estimated optical flow fields. Quantitative and comparative results are shown on the \emph{Middlebury} dataset~\cite{Middlebury} and a synthetic benchmark dataset with ground truth introduced by Garg \emph{et al.}~\cite{Garg}. Additionally, the performance of \emph{LCM-flow} is analyzed by varying several parameters such as the smoothness term weight and the vertex density of the input mesh. As \emph{LCM-flow} is proposed for non-rigid scenarios, it is natural to compare its result with non-rigid optical flow algorithms. The first non-rigid comparison is \emph{LCM-flow} against Garg~\emph{et al.}'s spatiotemporal optical flow approach which exploits correlations of neighboring pixels movement to constrain the flow computation. We also compare with the \emph{Improved TV-L1} (ITV-L1) algorithm and \emph{Large Displacement Optical Flow}~(LDOF). ITV-L1 preserves flow discontinuities by using a total variation regularization; and it ranks midfield of the \emph{Middlebury} evaluation. LDOF~\cite{LDOF} is proposed to overcome the large displacements issue by using region matching within a variational framework. It is also supposed as a baseline in relation to our \emph{LCM-flow} as those methods share a similar numerical minimization scheme (see Section~\ref{sec:Minimisation}). Our results show that the simple addition of our LCM constraint greatly improves accuracy and image interpolation. We also compare to the state of the art keypoint-based non-rigid image registration method proposed by Pizarro~\emph{et al.}~\cite{Pizarro}.

In our experiments we adopt a coarse-to-fine framework. An \emph{n}-level image pyramid is constructed for both input images, and each corresponding mesh. A down sampling factor of  0.75 is used on each pyramid level using \emph{Bicubic Interpolation}. The global smoothness constraint $E_{Global}$ is applied to each level of the pyramid with $\lambda$ set to 0.85. For minimizing the energy function on each level of the pyramid, the first fixed point iteration is set to 30 steps while the nested second fixed point iterations are fixed to 5 steps. Additionally, the large linear systems ~(Equations~\eqref{eq:EnergyKL_1} and \eqref{eq:EnergyKL_2}) are solved using \emph{Conjugate Gradients} with 45 iterations.

To summarize, our algorithm performs in the top tier of the \emph{Middlebury} interpolation criteria, and strongly overall - especially compared to the aforementioned specialist non-rigid optical flow techniques. In addition, we also demonstrate the strength of our approach in an application, i.e. constructing 3D Dynamic Morphable Models (3DDMMs)~\cite{ICCV}. We show our approach to outperform the proposed method of Cosker~\emph{et al.} in terms of accurate 3D mesh tracking across captured dynamic sequences.

\vspace{-3mm}

\subsection{Middlebury Dataset}

\begin{figure*}[t]
    \subfigure[Snapshot of \emph{Middlebury} optical flow evaluation (Submitted on February 2nd, 2012). Our proposed method is \emph{LCM-flow} with weight 0.6 and 25-pix mesh. In addition, the baseline methods are \emph{LDOF} and \emph{TV-L1-improved}~(Named ITV-L1 in context).]{\includegraphics[width=1\linewidth]{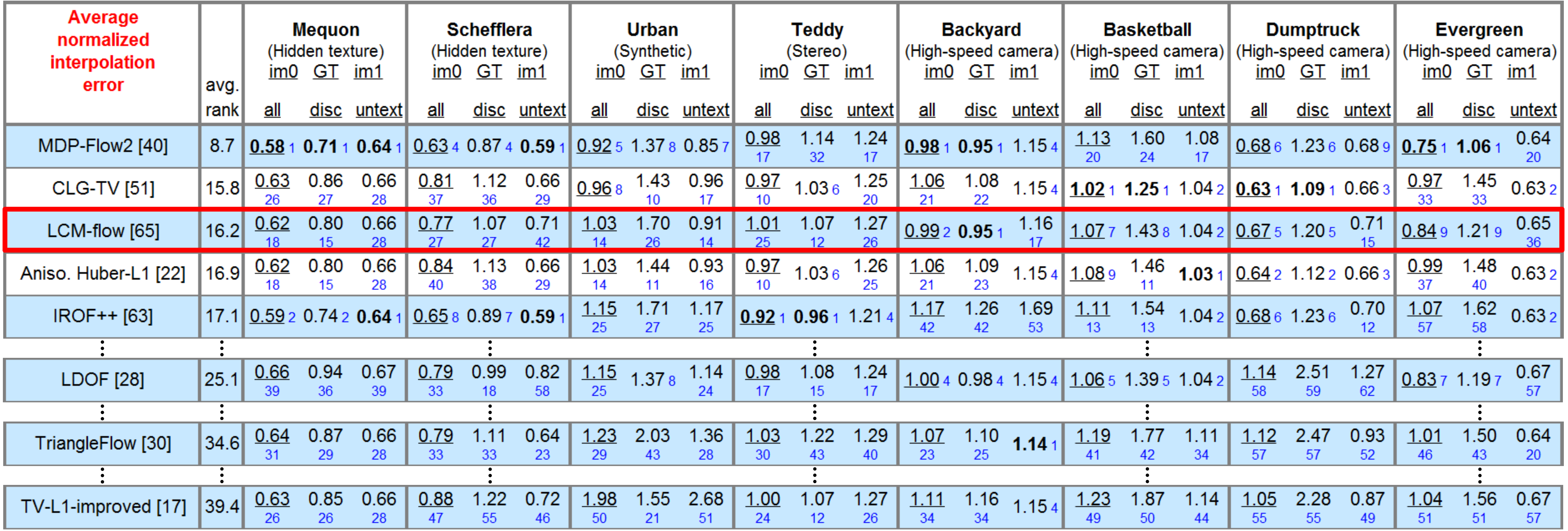}\label{fig:MuddleburyRanking}}\\
    \centerline{
    \captionsetup[subfigure]{labelformat=empty}
    \subfigure{\includegraphics[width=0.16\linewidth]{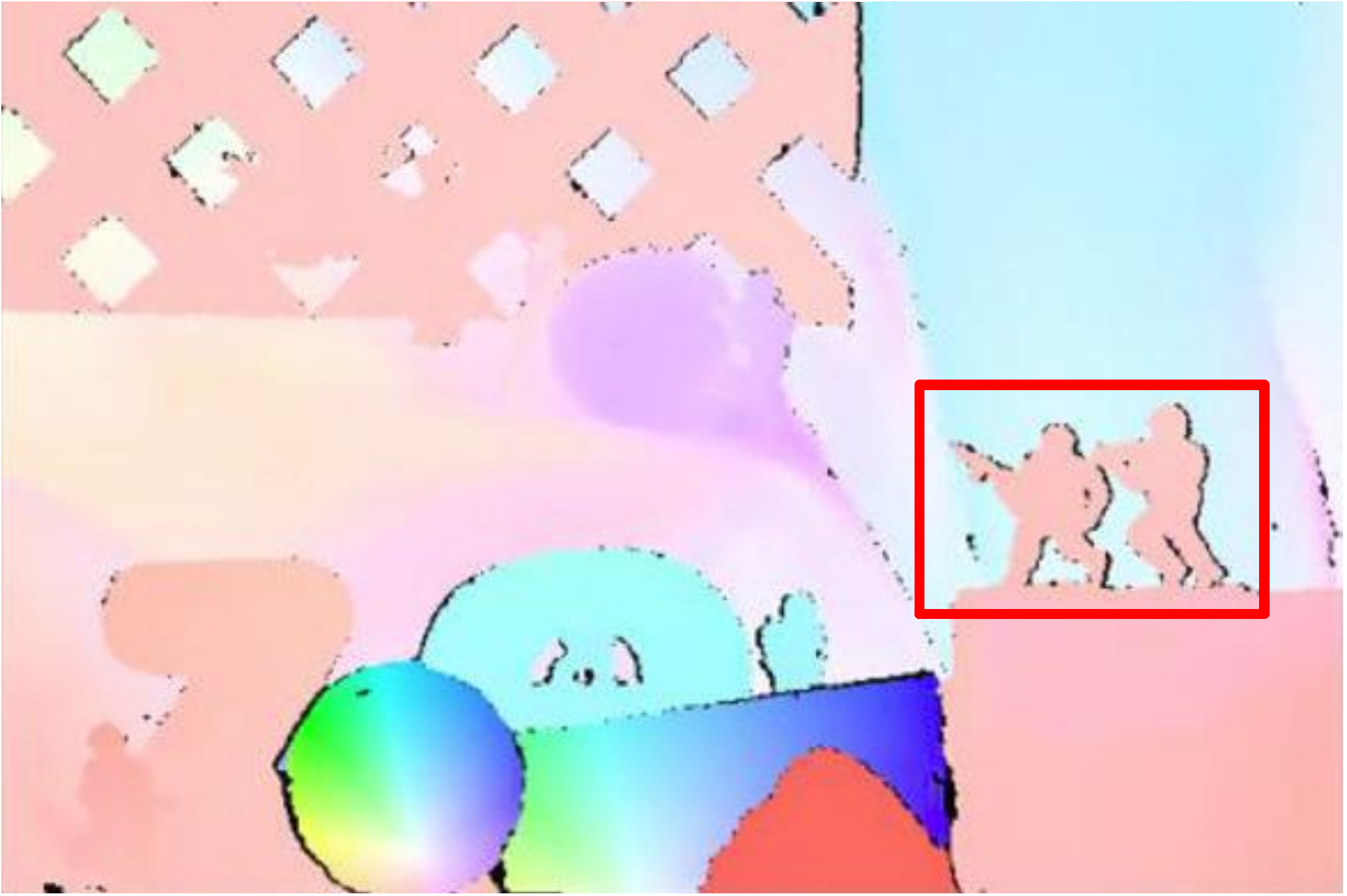}}\addtocounter{subfigure}{-1}
    \subfigure{\includegraphics[width=0.16\linewidth]{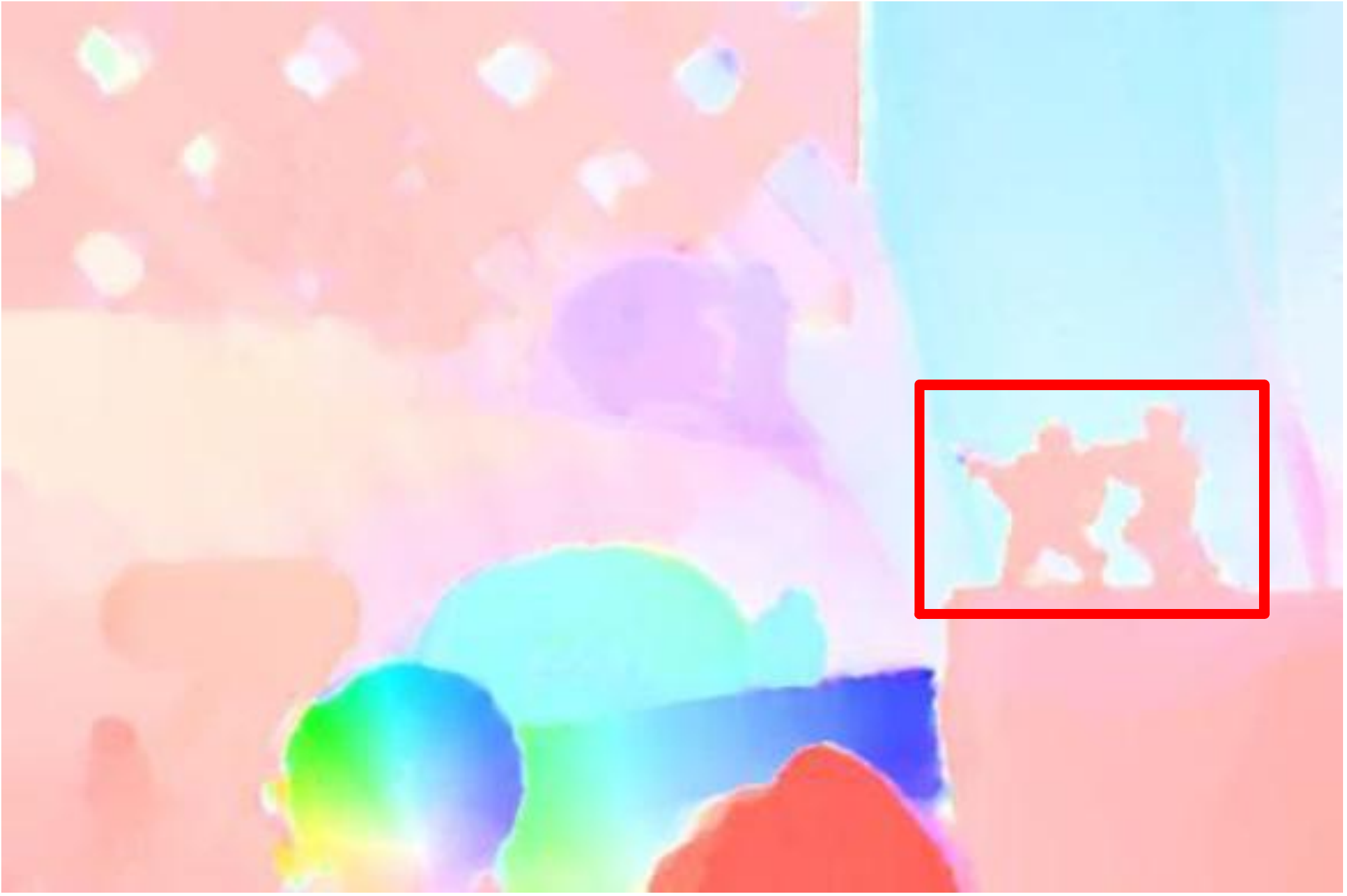}}\addtocounter{subfigure}{-1}
    \subfigure{\includegraphics[width=0.16\linewidth]{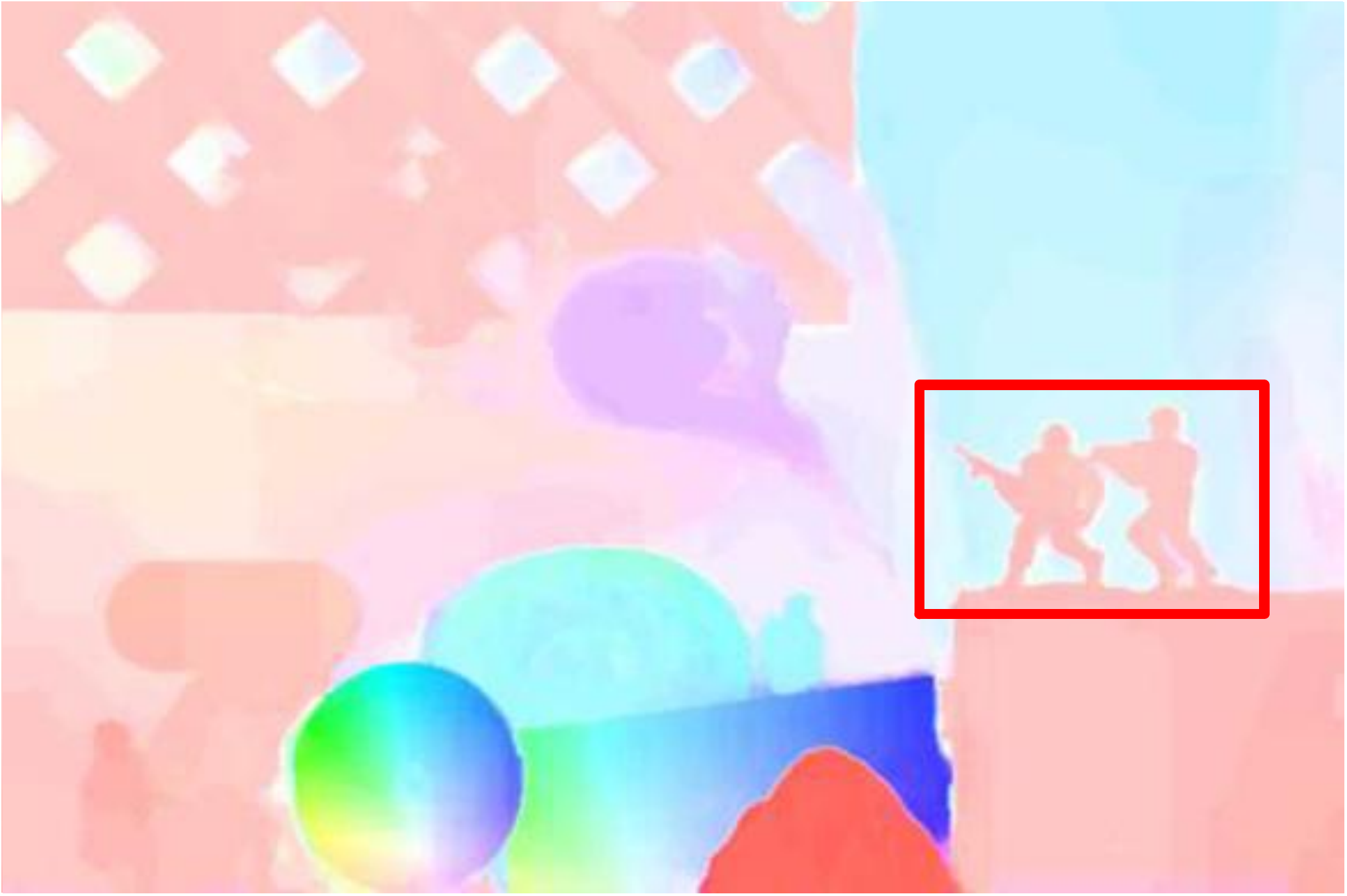}}\addtocounter{subfigure}{-1}
    \subfigure{\includegraphics[width=0.16\linewidth]{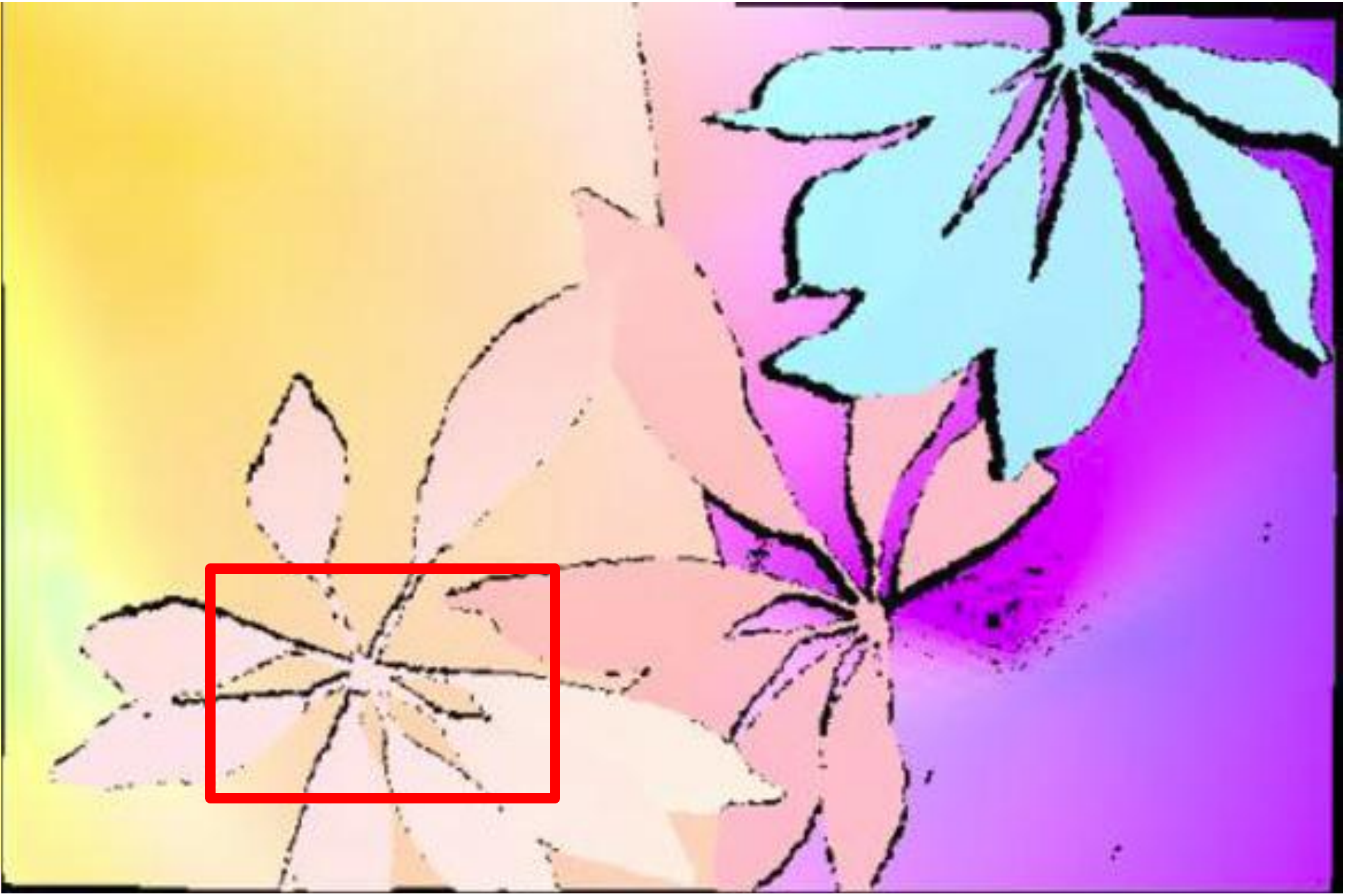}}\addtocounter{subfigure}{-1}
    \subfigure{\includegraphics[width=0.16\linewidth]{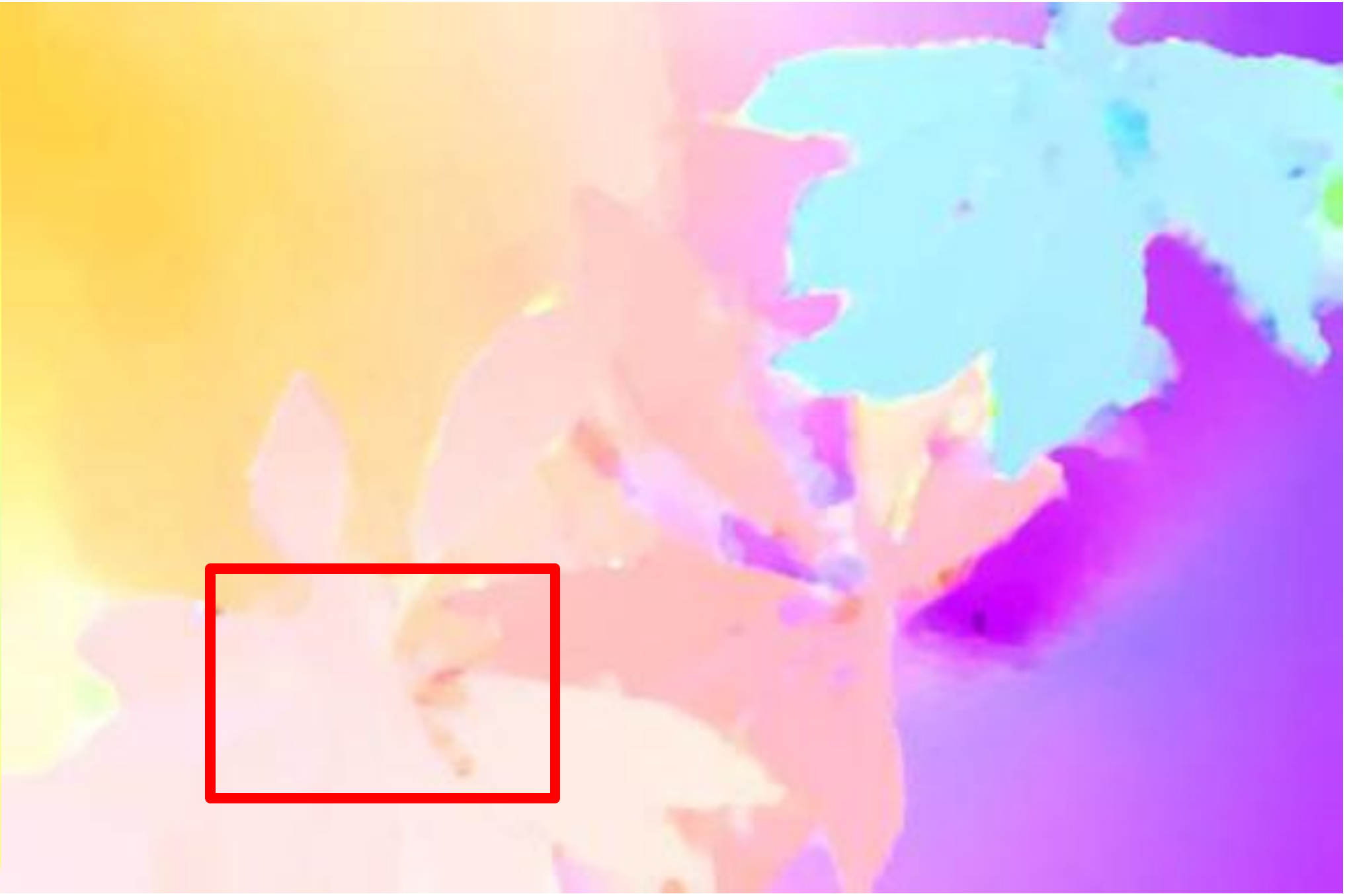}}\addtocounter{subfigure}{-1}
    \subfigure{\includegraphics[width=0.16\linewidth]{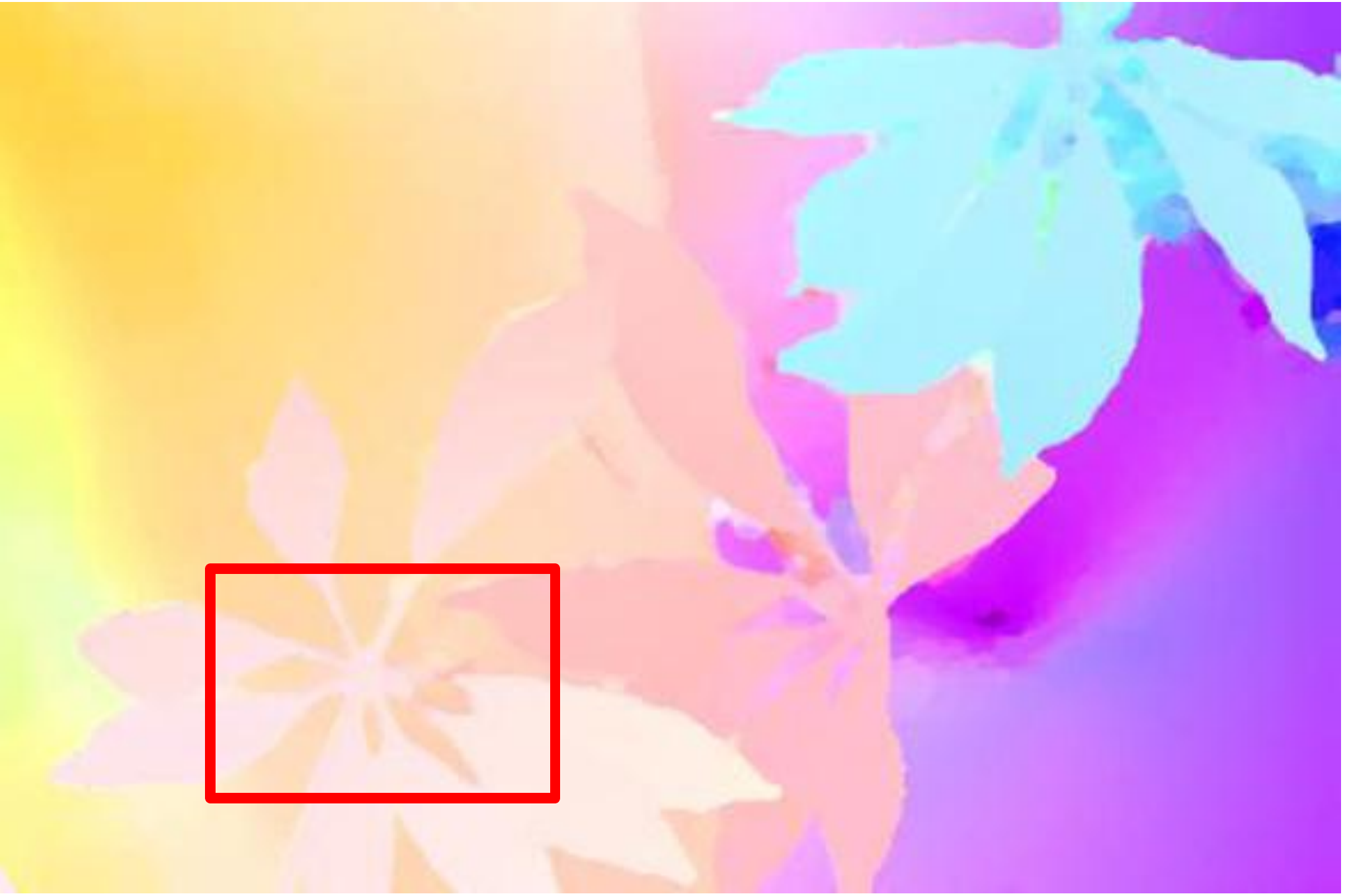}}\addtocounter{subfigure}{-1}
    }
    \centerline{
    \subfigure[]{\includegraphics[width=0.16\linewidth]{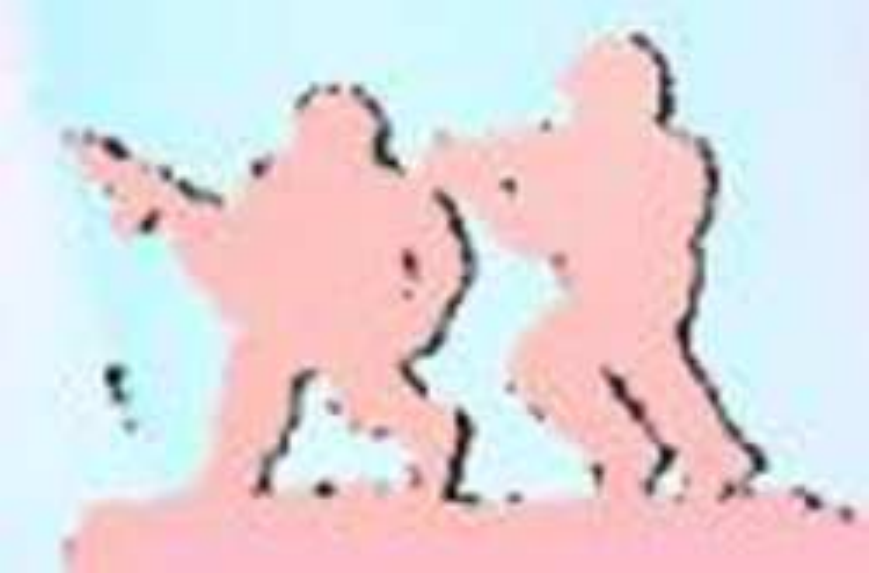}}
    \subfigure[]{\includegraphics[width=0.16\linewidth]{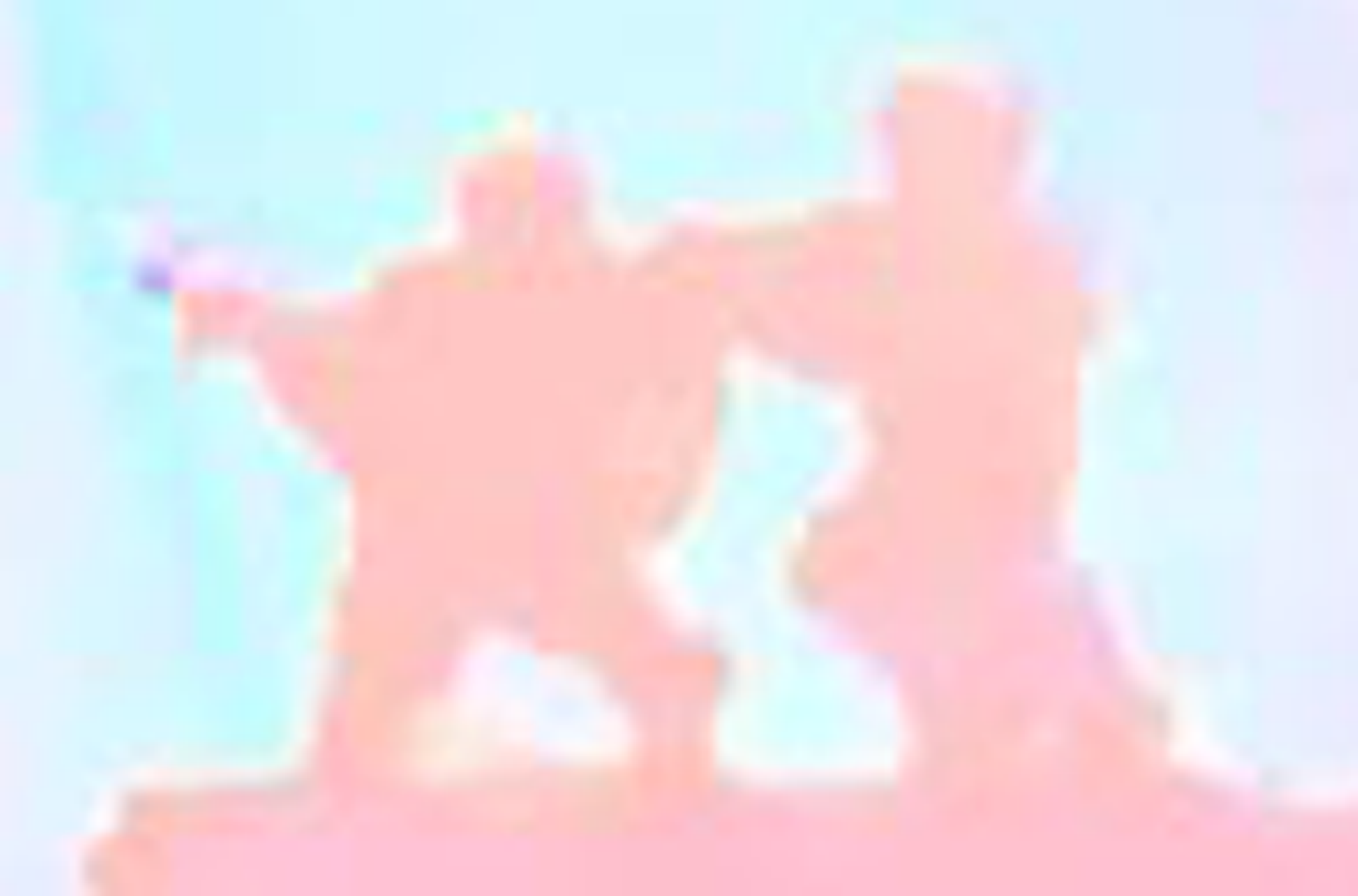}}
    \subfigure[]{\includegraphics[width=0.16\linewidth]{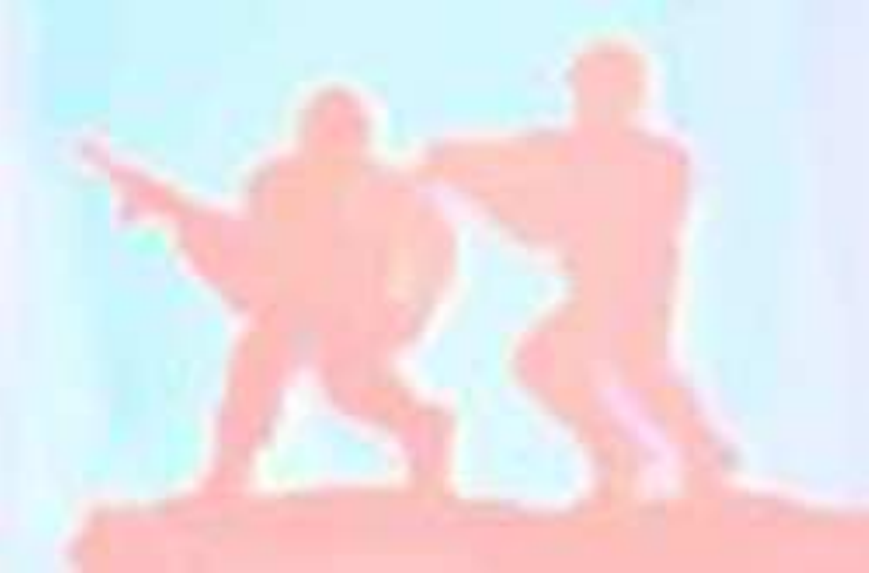}}
    \subfigure[]{\includegraphics[width=0.16\linewidth]{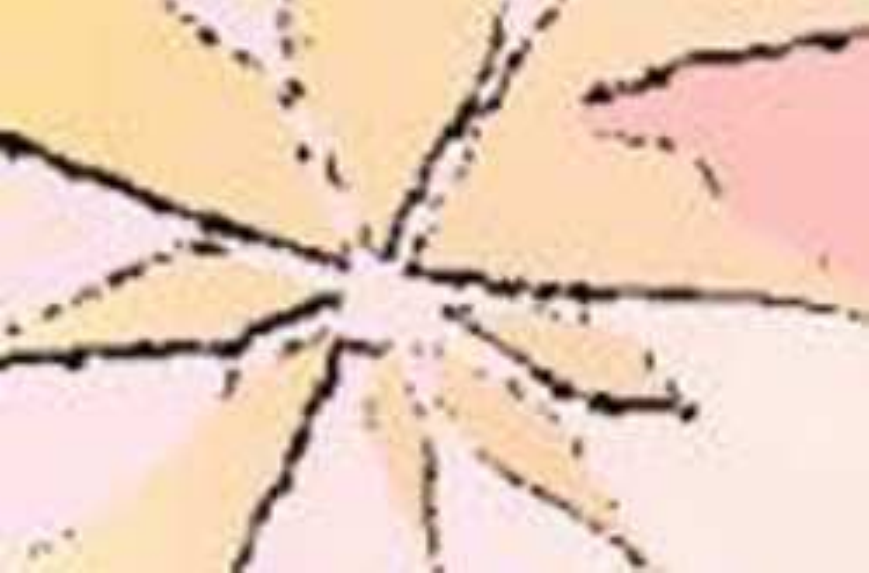}}
    \subfigure[]{\includegraphics[width=0.16\linewidth]{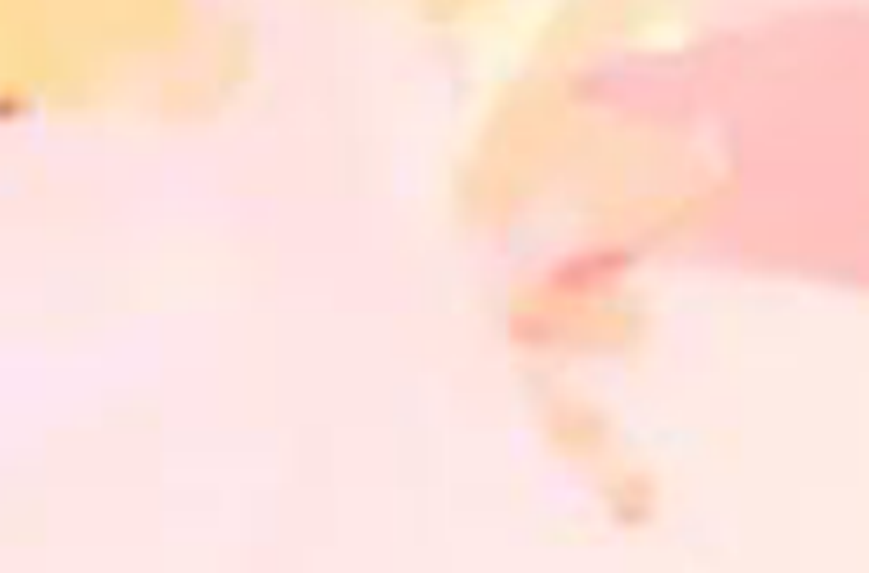}}
    \subfigure[]{\includegraphics[width=0.16\linewidth]{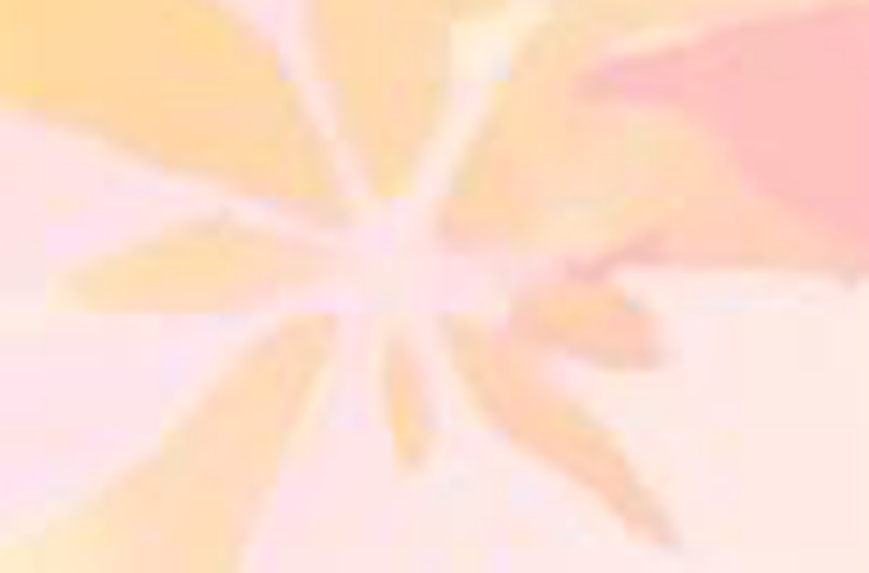}}
    }
    \caption{Screenshot of \emph{Middlebury} evaluation and visual comparison of optical flow fields calculated by \emph{LCM-flow} with different vertex densities for the input mesh on the \emph{Army} and \emph{Schefflera} sequences. (a) and (e): Ground truth optical flow field. (c) and (f): Weight 0.6 and 25-pix mesh. (d) and (g): Weight 0.6 and 10-pix mesh.}
    \label{fig:mbArmy}
\end{figure*}

We first performed an evaluation on the \emph{Middlebury} benchmark dataset. Two test cases of \emph{LCM-flow} are considered, each with different vertex densities for the input mesh. In the first test the proposed constraint weight is set as 0.6 and the vertex density of the mesh is fixed to 25 pixels - this being the distance between a vertex and its horizontal and vertical adjacent neighbors. The second test employs using a weight of 0.6 and a 10-pixel mesh.

As shown in Figure~\ref{fig:MuddleburyRanking}, \emph{LCM-flow} with a sparse (25-pix) input mesh ranks among the top three algorithms and significantly outperforms the baseline methods \emph{LDOF} and \emph{ITV-L1} in the \emph{Normalized Interpolation Error} test. Our overall average rank is also 16.2 compared to 25.1 and 39.4 respectively. \emph{Middlebury} results against other non-rigid approaches -- Garg~\emph{et al.}'s method and Pizarro~\emph{et al.}'s -- are not available. We therefore compare our approach against theirs using a different data set (see Section~\ref{sec:mocapDataset}). Our \emph{LCM-flow} approach also ranks fifth overall in the \emph{Interpolation Error} test. Particularly strong performance is observed on \emph{Middlebury} sequences captured using the high-speed camera -- \emph{Backyard}, \emph{Basketball}, \emph{Dumptruck} and \emph{Evergreen}.

In addition, \emph{LCM-flow} ranks in the reasonable midfield in both the \emph{Endpoint Error} and \emph{Angular Error} tests. This is again using the regular sparse mesh (25 pixels). Note that \emph{LCM-flow} has a lower ranking in this test, we believe this is due to the larger number of motion discontinuities which violate the local smoothness assumption of LCM-flow -- relative motion between foreground object and background often destroys the structure of the 1-ring neighborhoods crossed the boundary. The unrealistic deformation of 1-ring neighborhoods across the boundary often leads to the unexpected $\delta_{*}$ vector along the tangential direction of the boundary. One possible solution to this problem would be to use a denser input mesh. Figure~\ref{fig:mbArmy}(b-g) shows the visual effect of increasing mesh density, which results in a far sharper optical flow field on the boundary. Another possible solution would be to segment the scene and apply separate meshes for different objects, this is left for future work.

In the next section we compare against a recent popular optical flow data set specifically designed for non-rigid evaluation. In these trails we also perform further exploration of the effect of the various parameters in \emph{LCM-flow}.

\vspace{-3mm}

\subsection{MOCAP Benchmark Dataset}
\label{sec:mocapDataset}
\begin{figure*}[t]
    \centerline{
    \subfigure[Endpoint error comparison of different methods on Garg~\emph{et al.} benchmark dataset~\cite{Garg}.]{
    {\scriptsize
    \begin{tabular}{|c||cccc|cccc|}
    \hline
        & \multicolumn{4}{c|}{Average \bfseries{RMS Endpoint Error}}
        & \multicolumn{4}{c|}{\bfseries{$99^{th}$ percentile of Endpoint Error}} \\
        \bfseries{Methods}
        & \multicolumn{1}{c}{Original}
        & \multicolumn{1}{c}{Occlusion}
        & \multicolumn{1}{c}{Guass.N}
        & \multicolumn{1}{c|}{S\&P.N}
        & \multicolumn{1}{c}{Original}
        & \multicolumn{1}{c}{Occlusion}
        & \multicolumn{1}{c}{Guass.N}
        & \multicolumn{1}{c|}{S\&P.N}  \\ 
        \hline
        \textbf{\emph{Ours}, 0.8, 5-pix Mesh}  & 1.03 & 1.39 & 1.97 & \textbf{1.79} & \textbf{3.07} & 4.96 & \textbf{7.90} & \textbf{7.06}\\
        \textbf{\emph{Ours}, 0.8, 10-pix Mesh} & 1.16 & 1.51 & 2.07 & 1.86 & 3.33 & 5.28 & 7.99 & 7.28 \\
        \textbf{Garg \emph{et al.}, PCA~\cite{Garg}} & \textbf{0.98} & 1.33 & 2.28 & 1.84 & 3.08 & \textbf{4.92} & 8.33 & 7.09 \\
        \textbf{Garg \emph{et al.}, DCT~\cite{Garg}} & 1.06 & 1.72 & 2.78 & 2.29 & 6.70 & 5.18 & 7.92 & 8.53 \\
        \textbf{Pizarro \emph{et al.}~\cite{Pizarro}} & 1.24 & \textbf{1.27} & \textbf{1.94} & 1.79 & 4.88 & 5.05 & 8.67 & 8.54\\
        \textbf{ITV-L1~\cite{ITV_L1}} & 1.43 & 1.89 & 2.61 & 2.34 & 6.28 & 9.44 & 9.70 & 9.98\\
        \textbf{LDOF~\cite{Brox}} & 1.71 & 2.01 & 4.35 & 5.05 & 3.72 & 6.63 & 18.15 & 20.35\\
        \hline
    \end{tabular}
    \label{tab:comLCM}
    }
    }
    }
    \centerline{
    \subfigure[]{\includegraphics[width=0.22\linewidth]{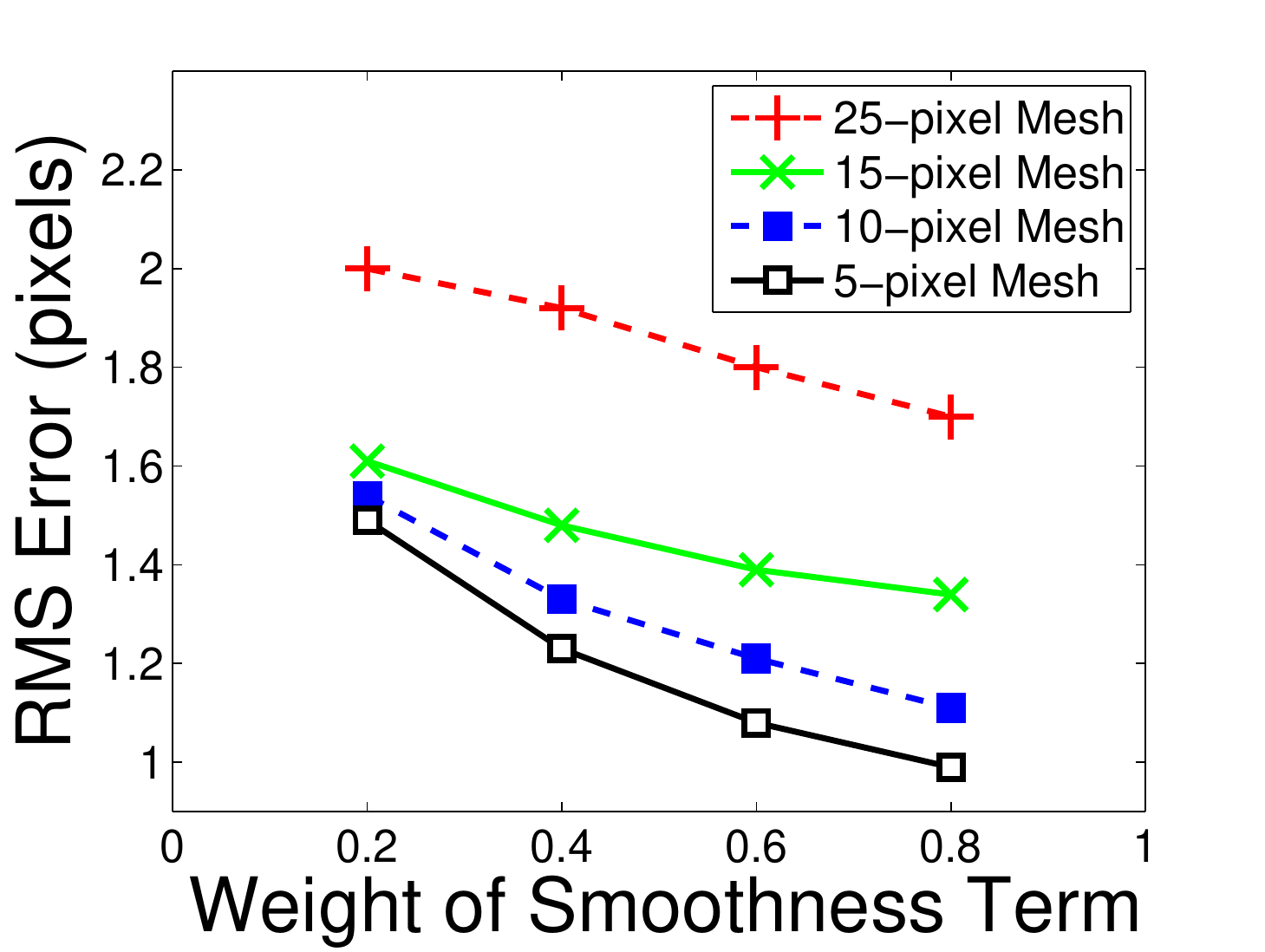}}
    \subfigure[]{\includegraphics[width=0.22\linewidth]{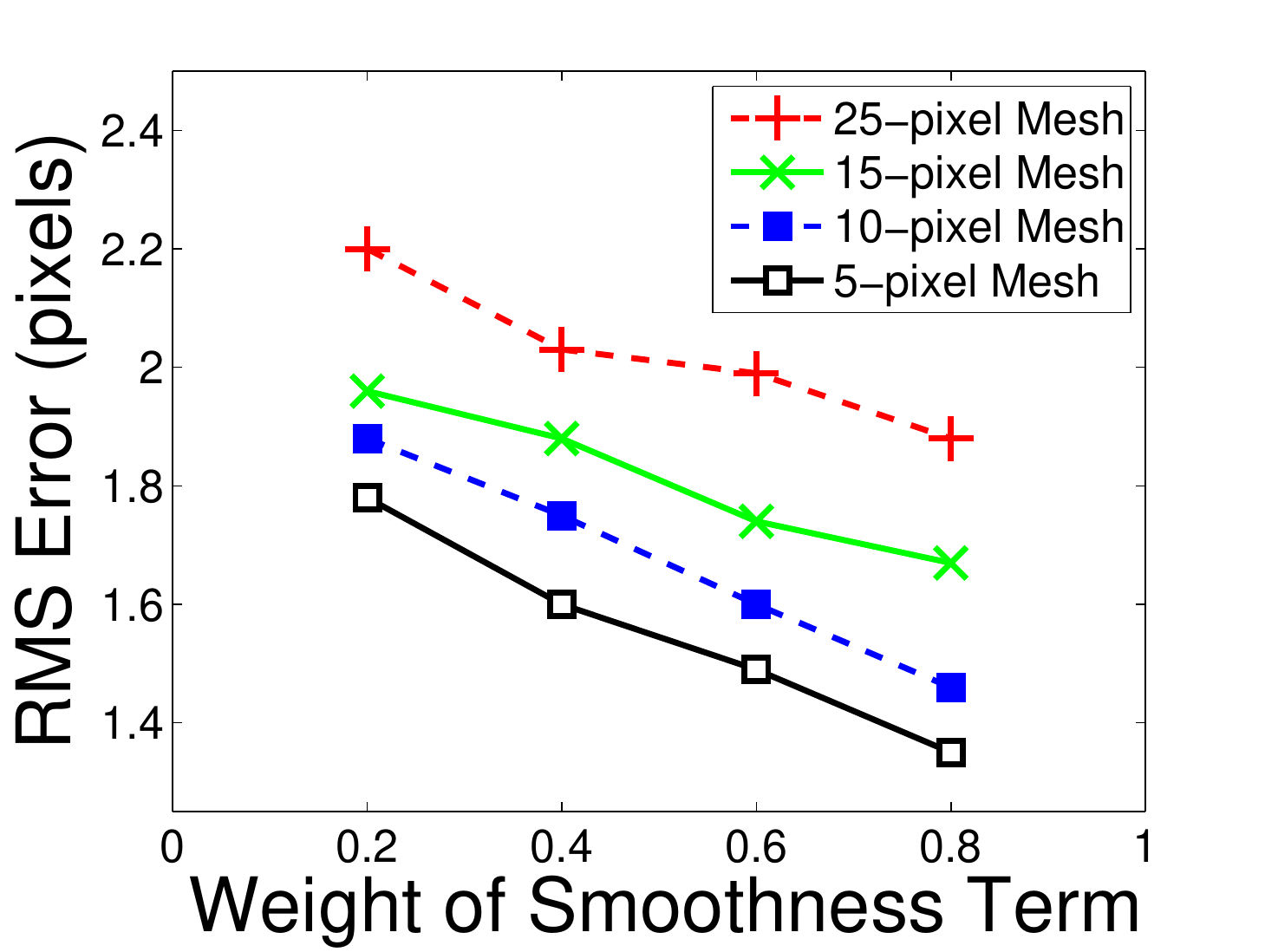}}
    \subfigure[]{\includegraphics[width=0.22\linewidth]{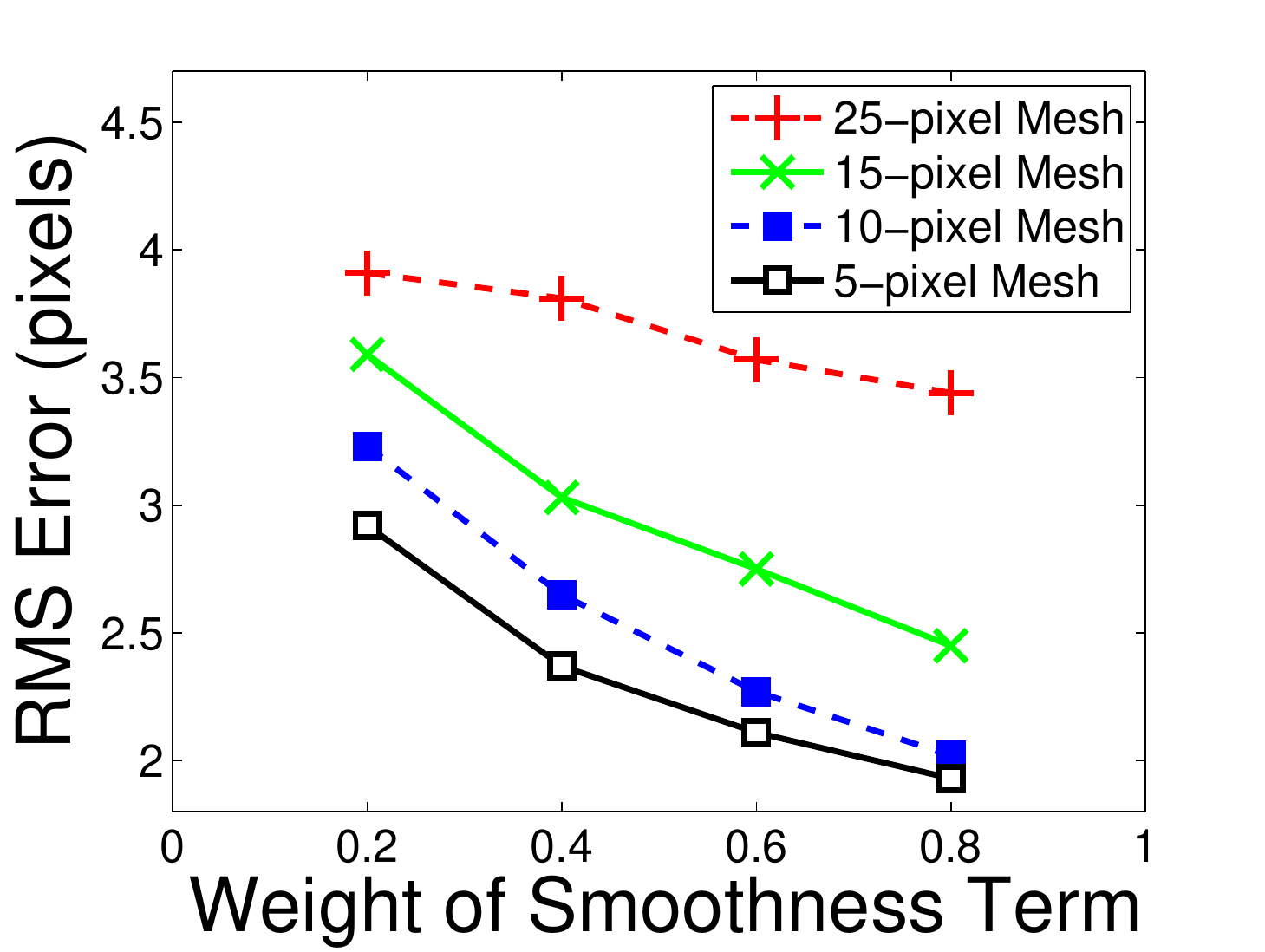}}
    \subfigure[]{\includegraphics[width=0.22\linewidth]{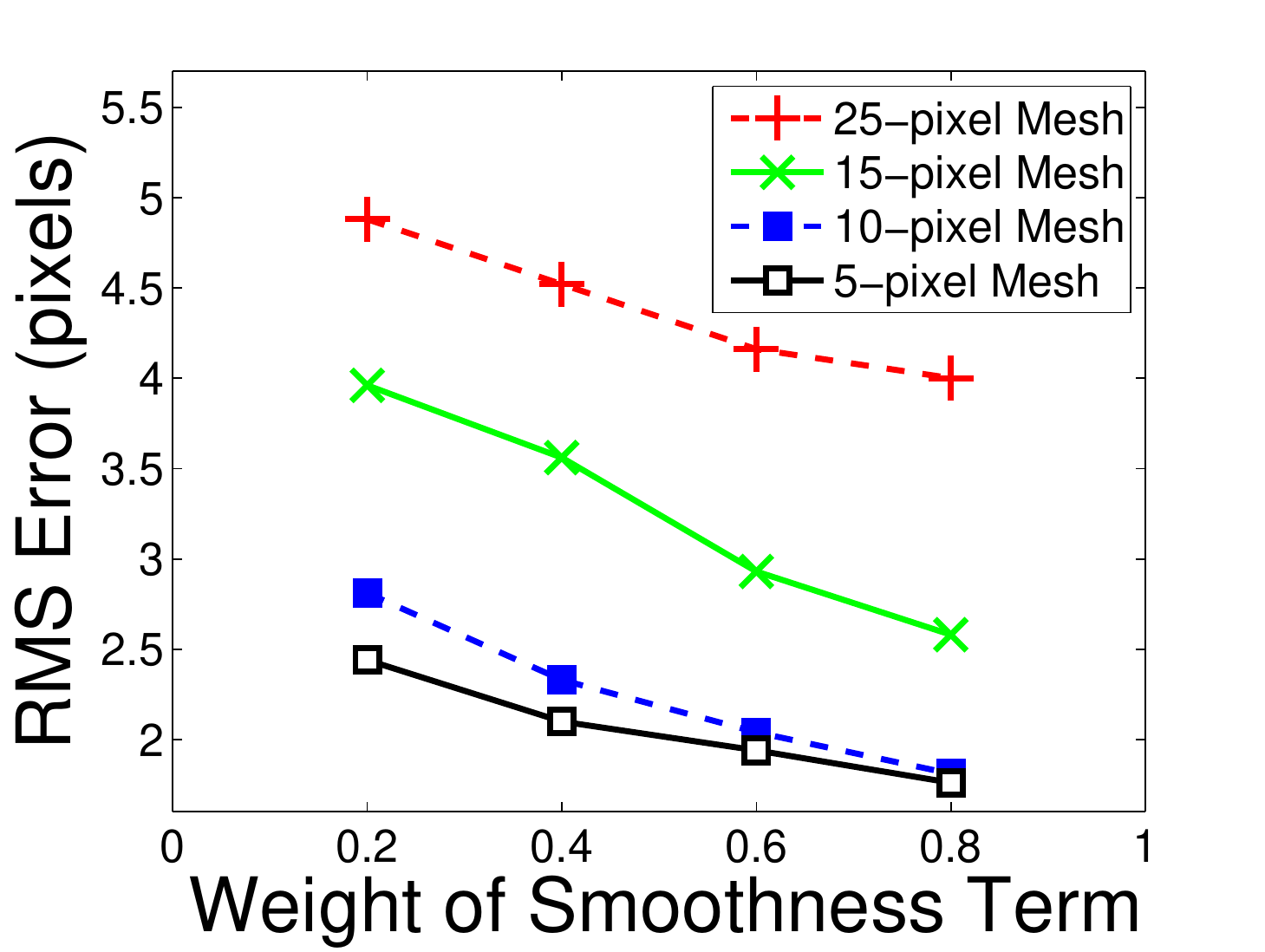}}
    }
    \vspace{-2mm}
    \caption{Result of \emph{LCM-flow} with different parameters on Garg~\emph{et al.} benchmark dataset. (a): Original. (b): Occlusion. (c): Gaussian noise. (d): S\&P noise.}
    \label{fig:selfCom}
\end{figure*}

To give a quantitative view on the non-rigid optical flow evaluation, Garg~\emph{et al.} introduces a benchmark sequence with dense ground truth flow fields~\cite{Garg}. To generate such a benchmark, they synthesise dense mesh by interpolating a sparse Vicon point based data from real deformable waving flag~\cite{white2007capturing}. Such a dense mesh is then projected onto the image texture plane, which gives a sequence (60 images, $500 \times 500$ pixel) along with dense ground truth flow field. To give more divergence on our evaluation, the experiments are performed on both this original GT sequence and three other degraded sequences, by adding three different noises:

\begin{itemize}
\item \textbf{Occlusions.} We give two black holes (20 pixels radius) that orbit the deformable surface.
\item \textbf{Gaussian noise.} We add Gaussian noise having 0.2 standard deviation relative to the range of image intensity.
\item \textbf{Salt \& pepper noise.} We add the salt and pepper with a 10\% density.
\end{itemize}

We evaluate the effect of varying both the weight of the \emph{LCM} smoothness term and the vertex density of the input mesh. The weight is varied with values of 0.2, 0.4, 0.6 and 0.8. The vertex density of the mesh is varied with values of 5 pixels, 10 pixels, 15 pixels and 25 pixels. As mentioned in the previous section, the same grid based triangulation is applied in all tests. Note that the weight 0 and no mesh tests are omitted because \emph{LCM-flow} degenerates to Brox~\emph{et al.} method~\cite{brox2004high} in this case, which is outperformed by our approach in the \emph{Middlebury} evaluation. When comparing against other methods, we use the same parameters cited by other authors. That is for both Garg~\emph{et al.} and ITV-L1 the weights $\alpha$ and $\beta$ are set to 30 and 2, we use 5 warp iterations, and 20 alternation iterations~\cite{ITV_L1}. According to parameters setting in~\cite{Garg}, the \emph{Principal Components Analysis} (PCA) and \emph{Discrete Cosine Transform} (DCT) are concerned for the 2D trajectory motion basis of Garg ~\emph{et al.} method in the context. In addition, the default parameters setting is used for LDOF~\cite{Brox}.

Figure~\ref{tab:comLCM} shows endpoint error (in pixel) comparisons on the four benchmark sequences of Garg~\emph{et al.}. \emph{LCM-flow} parameterized with a weight value of 0.8 and a 5 pixels mesh observes the best of \emph{Root Mean Square} (RMS) endpoint error on both \emph{noise} sequences and outperforms Garg~\emph{et al.} (DCT basis), ITV-L1 and LDOF algorithms on all four sequences. Garg~\emph{et al.} (PCA basis) has comparable performance (slightly outperforming us by 0.05 RMS) to our method on the original sequence, while in the occlusion sequence, Pizarro~\emph{et al.} leads overall. We also compute percentile-based accuracy measures~\cite{Seitz}. \emph{LCM-flow} parameterized with a weight of 0.8 and a 5 pixels mesh yields the best or second performance on all sequences except in the occlusion case where Pizarro~\emph{et al.} shows a marginal (0.12 RMS) improvement. As can be seen in Figure~\ref{fig:selfCom}, increasing the density of the input mesh results in significantly improved RMS errors in all trails. In a similar fashion, increasing the weight of \emph{LCM} smoothness term also yields more accurate optical flow estimation and reduced RMS error.

\begin{figure*}[t]
\centerline{
\includegraphics[width=0.95\linewidth]{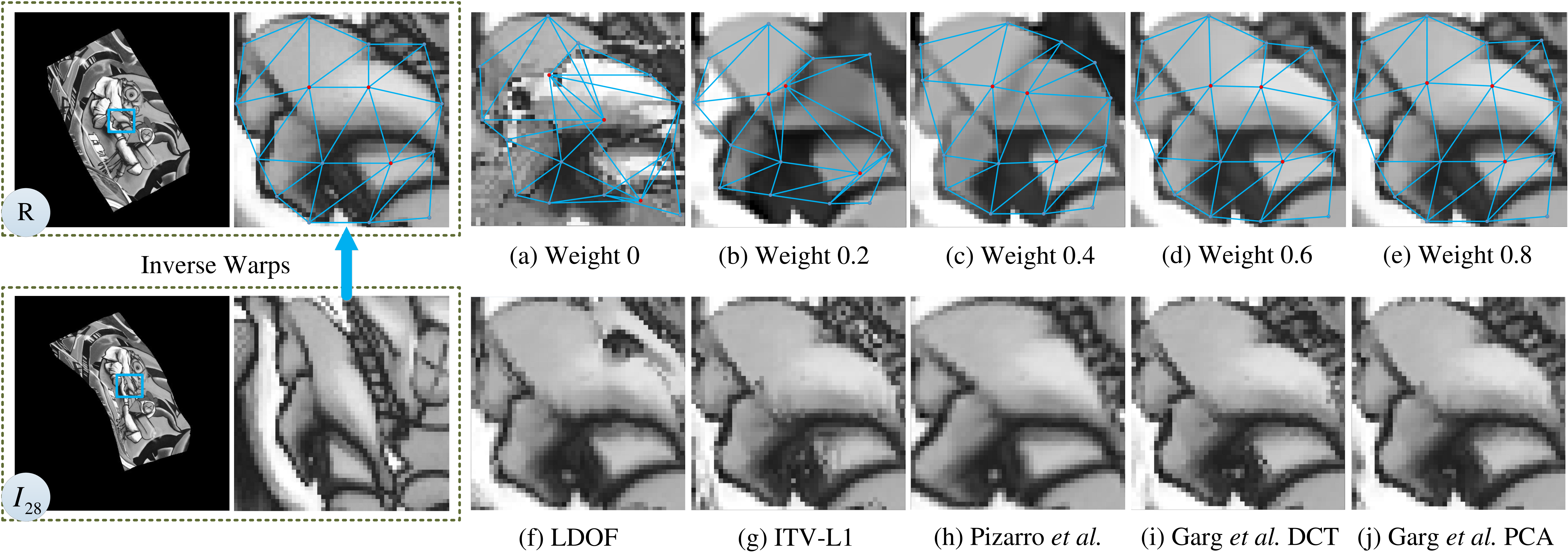}
}

\caption{Inverse warp details and triangle structure preservation. \textbf{Top-left box}: Reference frame and the initial mesh for \emph{LCM-flow}. \textbf{Bottom-left box}: Frame 28 of \emph{Original} sequence of Garg~\emph{et al.} benchmark dataset. (a): Alignment result and corresponding mesh of \emph{LCM-flow}, weight 0. (b): weight 0.2. (c): weight 0.4. (d): weight 0.6. (e): weight 0.8. (f): LDOF~\cite{Brox}. (g): ITV-L1~\cite{ITV_L1}. (h): Pizarro \emph{et al.}~\cite{Pizarro} (i): Garg \emph{et al.}, DCT basis~\cite{Garg}. (j): Garg \emph{et al.}, PCA basis~\cite{Garg}.}
\label{fig:triPer}
\end{figure*}

Figure~\ref{fig:triPer}(a-j) shows comparative \emph{Inverse Image Warping} results between \emph{LCM-flow} (with a non-uniform mesh and various weights) and five other state of the art algorithms on the Garg~\emph{et al.} \emph{Original} benchmark sequence. Examination of the images illustrates that increasing the \emph{LCM} constraint weight results in a sharper and less distorted image. This provides some insight into the algorithms strong performance in the \emph{Middlebury} interpolation result, as images warped using our computed flow appear to preserve local visual detail. In addition, Figure~\ref{fig:triPer}(a-e) shows the mesh corresponding to the input mesh on the reference frame (Top-left box). We observe that even applying a small weight to the \emph{LCM} smoothness term results in a stronger preservation of the triangular structure -- and hence a better preserved image structure.

\vspace{-3mm}

\subsection{Application: 3D Dynamic Morphable Model (3DDMM) Construction}

\begin{figure*}[t]
    \centerline{
    \subfigure[RMS values (in pixels) for tracked feature points versus ground truth landmark points.]{
    {\scriptsize
        \begin{tabular}{| c || c | c | c | c | c | c | c | c |}
        \hline
        \textbf{AU Seq.} & 20+23+25 & 9+10+25 & 16+10+25 & 1+2+4+5+20+25 & 18+25 & 12+10 & 4+7+17+23 & 1+4+15 \\
        \hline
        \textbf{AAM-TPS} & 53.8     & 25.4    & 23.2     & 18.6          & 15.3  & 15.2  & 3.4       & 2.3 \\
        \textbf{LCM-flow}& 2.0      & 3.0     & 1.1      & 1.7           & 1.8   & 1.5   & 2.0       & 1.4 \\
        \hline
        \end{tabular}
    }
    \label{tab:RMS}
    }
    }
    \centerline{
    \subfigure{\includegraphics[width=0.95\linewidth]{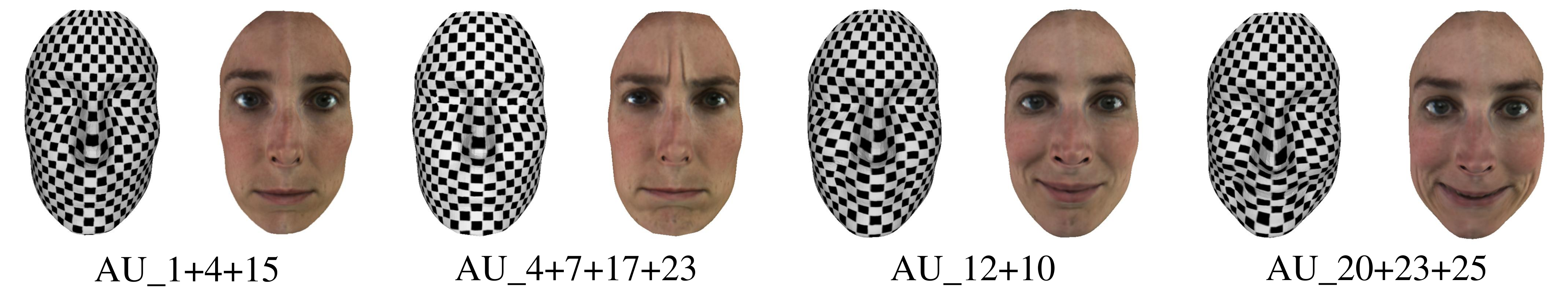}}
    }
\caption{Example output from a 3D Dynamic Morphable Model. The checkered pattern highlights correct underlying mesh deformation, which is dependent on accurate non-rigid UV map registration.}
\label{fig:selfCom}
\end{figure*}

In this part of the evaluation, we show the application of our algorithm to the construction of \emph{3D Dynamic Morphable Models}~\cite{ICCV}. These models are constructed from video-rate 3D facial scan data of different facial expressions. The essential problem with such data is aligning the 3D meshes such that all facial features are in correspondence. Solving this problem results in the same vertex topology deformed and tracked through the facial expression sequence. This can be approached by non-rigidly aligning the UV texture maps corresponding to the face meshes to a reference texture (e.g. a neutral expression), and then generating the 3D correspondences from these aligned images (see~\cite{ICCV} for full details). We applied our \emph{LCM-flow} algorithm to the alignment of the UV texture maps for 6 dynamic facial sequences and compared the results to those in Cosker~\emph{et al.} using the same ground truth labeled points. Table~\ref{tab:RMS} shows how the resulting the RMS errors for our \emph{LCM-flow} approach outperform each of the other methods, including the \emph{AAM-TPS} approach proposed by Cosker~\emph{et al.} After aligning the UV sequences we constructed a \emph{3D Morphable Model} from the corresponding meshes and rendered the output sequences. Figure~\ref{fig:selfCom} shows some example outputs, where a checkered pattern represents deformation in the underlying mesh.

\vspace{-3mm}

\section{Conclusion}
\label{conclusion}

\vspace{-2mm}

In this paper we have presented a novel optical flow formulation which proposes a \emph{Laplacian Cotangent Mesh} constraint for preserving local smoothness. Adapted from computer graphics, our term achieves this property by minimizes differentials in a Laplacian mesh. In our evaluation we have compared our method to several state of the art optical flow approaches on two well known evaluation sets. This has demonstrated our algorithms ability to provide accurate flow estimation and preserve local image detail -- evident through high scores in Middlebury evaluation, comparison to Garg \emph{et al.}, and experimentation on our algorithms parameters. In addition, we have demonstrated accurate results for in applying our algorithm for the construction of \emph{3D Dynamic Morphable Models}. For future work we are interested in more intelligently creating the underlying mesh to better approximate the image of interest. This should alleviate potential problems where triangles overlap the edges of multiple objects.

\section{Acknowledgement}

We thank Ravi Garg and Lourdes Agapito for providing their GT datasets and results. We also thank Gabriel Brostow and the UCL Vision Group for their helpful comments. The authors are supported by the EPSRC CDE EP/L016540/1 and CAMERA EP/M023281/1; and EPSRC projects EP/K023578/1 and EP/K02339X/1.

\section*{References}
\bibliographystyle{elsarticle-num}
\bibliography{cite}





\end{document}